\documentclass[runningheads]{llncs}
\usepackage[T1]{fontenc}
\usepackage{graphicx}
\usepackage{amsmath,amssymb,amsfonts}
\usepackage{algorithm}
\usepackage{algpseudocode}
\usepackage{booktabs}
\usepackage{tikz}
\usepackage{pgfplots}
\pgfplotsset{compat=1.18}
\usepgfplotslibrary{groupplots}
\usepackage{float} % http://ctan.org/pkg/float
\usepackage{multirow}

\usepackage{bm}
\begin{document}
\title{CALICO: Confident Active Learning \\with Integrated Calibration}
\titlerunning{CALICO: Confident Active Learning with Integrated Calibration}
\author{Lorenzo S. Querol\inst{1} \and
Hajime Nagahara\inst{1,2}\and Hideaki Hayashi\inst{1}}
\authorrunning{Querol et al.}
\institute{Osaka University, Japan \\
\email{\{lorenzoquerol\}@is.ids.osaka-u.ac.jp} \\
\email{\{nagahara,hayashi\}@ids.osaka-u.ac.jp} \and
Premium Research Institute for Human Metaverse Medicine (WPI-PRIMe)}

\maketitle              % typeset the header of the contribution

\begin{abstract}
The growing use of deep learning in safety-critical applications, such as medical imaging, has raised concerns about limited labeled data, where this demand is amplified as model complexity increases, posing hurdles for domain experts to annotate data. In response to this, active learning (AL) is used to efficiently train models with limited annotation costs. In the context of deep neural networks (DNNs), AL often uses confidence or probability outputs as a score for selecting the most informative samples. However, modern DNNs exhibit unreliable confidence outputs, making calibration essential. We propose an AL framework that self-calibrates the confidence used for sample selection during the training process, referred to as Confident Active Learning with Integrated CalibratiOn (CALICO). CALICO incorporates the joint training of a classifier and an energy-based model, instead of the standard softmax-based classifier. This approach allows for simultaneous estimation of the input data distribution and the class probabilities during training, improving calibration without needing an additional labeled dataset. Experimental results showcase improved classification performance compared to a softmax-based classifier with fewer labeled samples. Furthermore, the calibration stability of the model is observed to depend on the prior class distribution of the data.
\end{abstract}
\section{Introduction}
The growing complexity of modern deep neural networks (DNNs) poses a challenge by demanding a substantial increase in labeled data needed to achieve state-of-the-art performance~\cite{Gal2017_BayesianActiveLearning,Wu2022_surveyHumanInLoop}. In real-world applications, obtaining labeled data is a logistically expensive process. This challenge is particularly profound in medical imaging, where images are complex and difficult to interpret, requiring the need for domain experts with clinical experience. This implies that a longer turnaround time is needed to finalize ground-truth annotations \cite{Budd2021_MedicalImageAL,Wu2022_surveyHumanInLoop}. Given this time-consuming process, the ratio between labeled and unlabeled data samples becomes aggravated. To tackle this issue, methods have been devised to optimize data efficiency in training models.\par

Active learning (AL), or human-in-the-loop learning~\cite{Wu2022_surveyHumanInLoop}, is one of the existing methods that aims to reduce the need for extensive labeled data. The intuition behind AL is to involve human knowledge in the learning process by iteratively selecting samples that are considered most informative by some heuristic function \cite{Settles2009_ActiveLearningSurvey}. A subset of these samples is then given to a domain expert for annotation, thereby maximizing the performance of the model while concurrently minimizing the annotation costs.\par 

In AL, the confidence outputs of a DNN are commonly used to select the most informative samples~\cite{Wang2014_ActiveLabelingMethod}. In classification using DNNs, confidence is typically defined by the maximum value of the class posterior probabilities (i.e., $\max_c p(c \mid x)$ for class $c$ given an input sample $x$) calculated by the softmax function of the final layer. A lower confidence level indicates greater uncertainty in the model's prediction, making it more beneficial to label such samples. Consequently, these low-confidence samples are prioritized as the most informative for selection.\par

However, the straightforward method of using a softmax-based classifier was revealed to produce uncalibrated outputs~\cite{Grathwohl2019_ClassifierEnergyModel,Guo2017_calibrationModern,Ren2021_DeepALSurvey}. The problem arises from the characteristic of the cross-entropy loss function used in DNN training, where the loss decreases as the model's confidence approaches one. This happens when the posterior probability for a particular class approaches one while diminishing to zero for other classes. As a result, the classifier may erroneously exhibit high confidence even for input samples that are difficult to classify. This phenomenon is commonly referred to as the over-confidence issue. In the context of an AL paradigm, uncalibrated confidence outputs could impede reliable decision-making during the selection of informative samples. Consequently, this may lead to poor performance of the learned model on unseen data.\par

Hence, a concept called \textit{confidence calibration} in neural networks becomes a crucial aspect of developing modern intelligent systems. Confidence calibration is defined as the reflection of a model's accuracy with its predictive confidence~\cite{Guo2017_calibrationModern,minderer2021_revisitCalibration}. For instance, in scenarios where a classifier provides 100 predictions, each with 95\% confidence, it is statistically expected that 95 of those predictions should be correct. In general, the calibration of neural networks is performed in a post-hoc manner, which calibrates the confidence output of the model after training. A common method for this is temperature scaling, which modifies the softmax function in the final layer by incorporating a temperature parameter, thereby calibrating the confidence output. However, post-hoc methods typically require a separate labeled dataset, inefficient in the context of AL as it consumes the already limited labeled data.\par

\begin{figure}[t]
    \centering
    \includegraphics[width=0.9\linewidth]{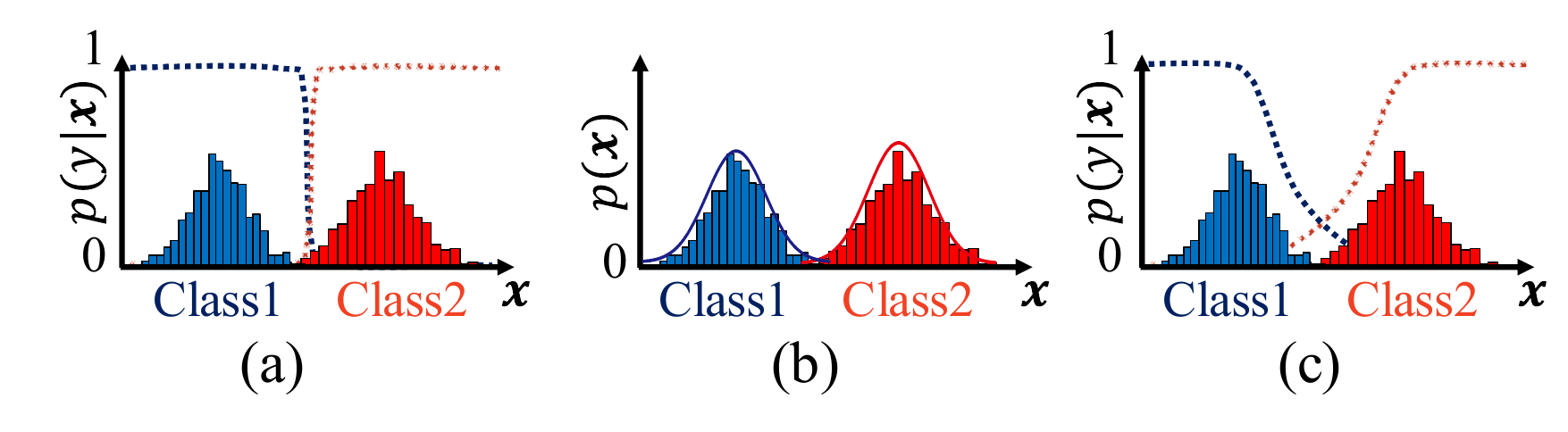}
    \caption{Joint learning of a classifier and generative model and its advantage. (a) Training solely a classifier can lead to inaccurately high posterior probability near the class decision boundary. (b) Estimating the input distribution using a generative model allows us to account for the frequency of data occurrences. (c) Joint learning of the classifier and generative model helps to calibrate the confidence scores.}
    \label{fig:joint_learning_benefit}
\end{figure}

To calibrate the confidence output during the AL loop without relying on validation samples, our key idea involves leveraging the distribution of unlabeled training data. This is achieved through the simultaneous learning of a classifier and generative model. Fig.~\ref{fig:joint_learning_benefit} outlines the advantages of this joint learning approach. Training a classifier in isolation often leads to inaccurately high posterior probability near the decision boundary. However, by concurrently estimating the input data distribution with a generative model, the classifier is trained to account for the frequency of data occurrences. This approach naturally lowers the posterior probabilities for ambiguous data points near the decision boundary, leading to an inherent self-calibration of confidence levels.\par
% ====================================
% Fig Intuitive explanation of the benefit of joint learning
% ====================================
% ====================================

We propose an AL framework designed to self-calibrate confidence during the training process. The method, called CALICO (\textbf{\underline{C}}onfident \textbf{\underline{A}}ctive \textbf{\underline{L}}earning with \textbf{\underline{I}}ntegrated \textbf{\underline{C}}alibrati\textbf{\underline{O}}n), involves the joint training of a neural network-based classifier with an energy-based model (EBM)~\cite{LeCun2006_EnergyBasedLearning} in a semi-supervised manner. This joint training enhances the model's understanding of the input data distribution, thereby calibrating the confidence outputs. The key idea is to use these calibrated confidence outputs as input for a query strategy, specifically using the least confidence strategy. This approach enhances decision reliability in selecting samples for annotation, with the overall goal of minimizing model miscalibration and improving accuracy with a minimal number of samples.\par

The contributions of this paper are as follows:
\begin{enumerate}
    \item We propose an AL framework termed CALICO, which is designed to self-calibrate confidence during the training process. Our method involves joint training of a classifier and EBM to achieve calibration of confidence without separate validation samples, utilizing the calibrated confidence outputs to select the most informative samples.
    \item We demonstrate that the self-calibration approach, which involves simultaneous learning of a classifier and a generative model, is effective for AL in terms of improving accuracy and decreasing calibration error using fewer labeled data than straightforward baseline methods.
    \item We revealed the potential of class distribution balancing to enhance CALICO's performance on datasets with class imbalance.
\end{enumerate}

\section{Background \& Related Works}
\subsection{Active Learning}
\begin{figure}[t]
\centering\includegraphics[width=0.9\textwidth]{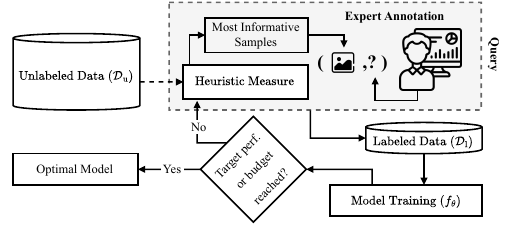}
    \caption{An overview of the typical active learning (AL) cycle.} 
    \label{fig:al-cycle}
\end{figure}
According to \cite{Settles2009_ActiveLearningSurvey}, allowing a learnable algorithm to select the data from which it learns can lead to more effective learning. In other words, AL seeks to maximize neural network performance with a smaller selection of data. As shown in Fig. \ref{fig:al-cycle}, a typical AL scenario involves an \textit{active learner} (model) continuously seeking new samples from a large pool of unlabeled data and inquiring the \textit{oracle} (domain expert) for ground-truth annotations. The goal is to achieve a specific target performance, such as accuracy, while minimizing labeling costs. AL has been widely utilized in traditional machine learning tasks \cite{Settles2009_ActiveLearningSurvey,Fu2012_InstanceSelection,Kumar2020_ALQueryStrategies}, but with the emerging use of DL achieving superior results in various tasks \cite{Wang2014_ActiveLabelingMethod,Gal2017_BayesianActiveLearning} comes with increasing dependence on the amount of labeled data gathered. Acquiring labeled data in task-specific or real-world problems is laborious and time-consuming, thus methods such as AL have become more relevant with its use in combination with DL, hence commonly referred to as deep AL \cite{Zhan2022_ActiveLearningSurvey}. \par

Sample acquisition can be categorized into three main frameworks: membership query synthesis, stream-based sampling, and pool-based sampling. In real-world scenarios, it is common to acquire a large batch of unlabeled data at once, prompting the use of pool-based sampling \cite{Settles2009_ActiveLearningSurvey}. The pool-based sampling framework assumes that there is a limited amount of labeled data and a more extensive pool of unlabeled data. To guide the active learner in selecting which data points to request labels for, samples are selected greedily via a \textit{query strategy}~\cite{Wang2014_ActiveLabelingMethod}. The labels for the query are obtained by inquiring the oracle, and the data pools are consequently updated. The model is then retrained iteratively until a certain metric reaches a target value or the unlabeled data pool is exhausted. Various common query strategies, such as maximum entropy, margin, least confidence, and mean standard deviation-based approaches, have been established in AL. Moreover, the rise of frameworks like generative adversarial networks and Bayesian deep learning \cite{Gal2015_BayesCNN,Tran2019_BayesGenerative} have also contributed to the development of enhanced query strategies in this field. \par

Despite extensive research on query strategies in AL, extending these methods to DL remains challenging, especially due to model uncertainty and inadequate labeled data \cite{Zhan2022_ActiveLearningSurvey}. The often utilized softmax response for deriving the class probability in DNNs exhibits overconfidence~\cite{Wang2017_CostEffectiveAL}, posing possible difficulties in evaluating unlabeled data. This observation has attracted research attention towards deep Bayesian AL~\cite{Gal2015_BayesCNN,Gal2017_BayesianActiveLearning,Tran2019_BayesGenerative}. Furthermore, recent advances in uncertainty quantification have received interest for its use in AL paradigms~\cite{abdar2021_reviewUQ,hein2022_comparisonUQ}. However, these strategies frequently result in increased computation time for model training and inference, and may require altering the model architecture itself. To solve the dilemma of limited labeled data in deep learning, an approach involves reducing the issue by merging semi-supervised learning with AL~\cite{Simeoni2019_SSLActive}. This strategy uses unlabeled data to gather more information about the data distribution and improves the model's overall performance. \par

% The research direction of DAL has evolved toward task-specific or domain-specific applications~\cite{Zhan2022_ActiveLearningSurvey}, including areas such as out-of-distribution detection \cite{kothawade2021_activeOOD}, addressing rare or imbalanced classes~\cite{aggarwal2020_activeImbalanced}, and handling large-scale unlabeled data pools~\cite{citovsky2021_activeScale}. While these developments demonstrate a tailored approach to specific challenges, there is still room for more experimentation and interesting advancements in the domain of DAL. With the increasing use of deeper and wider networks due to their favorable performance in a wide range of tasks~\cite{Guo2017_calibrationModern}, modern classifier architectures have become a common use-case to develop intelligent systems. 

\subsection{Energy-Based Models}
An EBM is a type of generative model that directly models a negative log-probability, which is also known as the energy function. The model's probability density function is derived by normalizing the energy function, expressed as $p_{\bm{\theta}}(\bm{x}) = \exp(-E_{\bm{\theta}}(\bm{x}))/Z_{\bm{\theta}}$. In this formulation, $E_{\bm{\theta}}$ is the energy function parameterized by $\bm{\theta}$, and $Z_{\bm{\theta}}$ is the normalizing constant, computed as $Z_{\bm{\theta}} = \int_{\bm{x}} \exp(-E_{\bm{\theta}}(\bm{x})) \mathrm{d}\bm{x}$. A key challenge in working with EBMs is that this integral for the normalizing term is typically intractable. To estimate it, advanced sampling techniques are often employed, such as using Markov chain Monte Carlo (MCMC) and stochastic gradient Langevin dynamics (SGLD)~\cite{welling2011bayesian}. \par

EBMs are utilized in a vast range of applications such as image generation~\cite{du2019implicit}, texture generation~\cite{xie2018cooperative}, and text generation~\cite{deng2020residual,bakhtin2021residual}. EBMs have also applied in the context of dropout and pruning within NNs~\cite{salehinejad2021edropout}. Additionally, EBMs can be utilized for the complex problem of continuous inverse optimal control~\cite{Xu2022energy}. There have been several approaches to simultaneously training an EBM and a classifier~\cite{Yang2021_JemPlusPlus,Liu2022_ssmJEM,Yang2023_sadaJEM,Hayashi2023_gaussianCoupled}, revealing the effectiveness of EBMs for semi-supervised learning, outlier detection, and confidence calibration. \par 

\subsection{Confidence Calibration} \label{sec:conf-calib}
Formally, considering a dataset with input $x$ and outcome $y\in \{1,\ldots, K\}$, a neural network is considered \textit{perfectly calibrated} if:
\begin{equation}
p(Y=y\mid\hat{c}=c)=c, \quad \forall c\in[0,1]
\end{equation}
\noindent Here, $\hat{c}$ represents the probability of a predicted label $Y$, and $y$ is the ground-truth label. Model calibration is frequently depicted using reliability diagrams \cite{Guo2017_calibrationModern}, where the expected accuracy is plotted as a function of confidence. Perfect calibration is plotted similarly as an identity function, and any deviation from the diagonal signifies miscalibration. 

% This approach allows them to extract contextual information from the distribution space and learn correct labels from the formed clusters.

% the instability observed during training initiated a series of developments aimed at incorporating new techniques to enhance training speed and stability \cite{Yang2021_JemPlusPlus,Liu2022_ssmJEM}, fine-tuned control between discriminative or generative performance \cite{Hayashi2023_gaussianCoupled}, and general improvements in discriminative performance and image sample quality \cite{Yang2023_sadaJEM}. 

Literature reveals that the utilization of modern neural networks with increasing complexity often results in poor calibration \cite{Guo2017_calibrationModern,minderer2021_revisitCalibration}. Therefore, various methods have been proposed to calibrate confidence. The calibration methods can be classified into two categories: post-hoc calibration and train-time calibration. Post-hoc calibration is one performed after training. It has been a primal approach, and many methods have been proposed~\cite{Guo2017_calibrationModern}. Among such methods, temperature scaling~\cite{Guo2017_calibrationModern} is considered simple and effective. It uses a softmax function with temperature instead of the usual softmax in the final layer of the NN and tunes the temperature parameter to optimize the negative likelihood. In contrast, train-time calibration is performed during training, which involves simultaneous learning with generative models~\cite{Grathwohl2019_ClassifierEnergyModel,Hayashi2023_gaussianCoupled} and the use of soft labels~\cite{thulasidasan2019mixup-calib}. The main difference between post-hoc and train-time methods is whether the validation data is used to calibrate the confidence.

\section{CALICO: Confident Active Learning with Integrated Calibration}
The proposed CALICO achieves efficient learning by utilizing calibrated confidence when selecting samples to be labeled. Confidence calibration is performed during the AL process by simultaneously estimating the input data distribution with an EBM and the class posterior probabilities with a classifier. 
\subsection{Algorithm of CALICO}
\begin{algorithm}[t]
\caption{CALICO: Confident Active Learning with Integrated Calibration}
\label{alg:calib-conf}
\begin{algorithmic}[1] 
\State \textbf{Input:} Labeled dataset $\mathcal{D}_\mathrm{l}$, Unlabeled pool $\mathcal{D}_\mathrm{u}$, Query Size $Q$, Number of queries $N_\mathrm{q}$, Query strategy $\alpha^{\mathrm{LC}}$, Oracle $g$
\State \textbf{Initialize:} Model $f_{\bm{\theta}}$
% \State $H \gets \alpha^{\mathrm{LC}}(x, f_{\theta})$ \Comment{Heuristic function}
% \State $g \gets \textbf{R}^D \in \{\mathrm{C_1},\ldots,\mathrm{C_K}\}$ \Comment{Oracle}
\While{$1 \leq i \leq N_\mathrm{q}$}
    \State \textbf{Train:} Train $f_{\bm{\theta}}$ over $\mathcal{D}_\mathrm{l} \cup \mathcal{D}_\mathrm{u}$
    % \For{$1 \leq j \leq Q$} 
    \State $\mathcal{D}_\mathrm{q} \gets \alpha^{\mathrm{LC}}(\mathcal{D}_\mathrm{u}, Q, f_{\bm{\theta}})$ %\Comment{Run $\alpha^{\mathrm{LC}}$ over $\mathcal{D}_\mathrm{u}$ to get $\mathcal{D}^*_{q}$}
    % \EndFor
    % \State \# Get $Q$ most informative samples
    % \For{$1 \leq k \leq Q$} \
    \State $\mathcal{D}_\mathrm{l} \gets \mathcal{D}_\mathrm{l} \cup g(\mathcal{D}_\mathrm{q})$  
    \State $\mathcal{D}_\mathrm{u} \gets \mathcal{D}_\mathrm{u}\setminus \mathcal{D}_\mathrm{q}$ 
    % \EndFor
    \State \textbf{Evaluate:} Compute performance of $f_{\bm{\theta}}$ on a valid. set
    % \State $i \gets i + 1$ \Comment{Inc. iteration counter}
\EndWhile
\State \textbf{Output:} Trained model $f_{\bm{\theta}}$
\end{algorithmic}
\end{algorithm}

The details of CALICO are described in Algorithm~\ref{alg:calib-conf}. Suppose that we have a limited amount of labeled data $\mathcal{D}_\mathrm{l}=\{(\bm{x}_i,y_i)\}^{M}_{i=1}$, and a more extensive pool of unlabeled data $\mathcal{D}_\mathrm{u}=\{\bm{x}_i\}^{N}_{i=1}$, where $M<N$, and $y_i\in\{1,\ldots, K\}$ is the class label of input $\bm{x}_i$. The model $f_{\bm{\theta}}$ is trained over both $\mathcal{D}_\mathrm{l}$ and 
$\mathcal{D}_\mathrm{u}$. The use of unlabeled data is enabled by simultaneous learning with an EBM, and this point differs from traditional AL processes. The query $\mathcal{D}_\mathrm{q}$ is a subset of $\mathcal{D}_\mathrm{u}$ and is selected based on a predefined query size $Q$ and query strategy $\alpha^{\mathrm{LC}}$, which is 
detailed in the next subsection. The labels for $\mathcal{D}_\mathrm{q}$ are obtained by inquiring with the oracle $g$, and $\mathcal{D}_\mathrm{q}$ with the obtained labels is denoted as $g(\mathcal{D}_\mathrm{q})$. Subsequently, both data pools are updated accordingly where $\mathcal{D}_\mathrm{l} \leftarrow  \mathcal{D}_\mathrm{l} \cup 
g(\mathcal{D}_\mathrm{q})$, and $\mathcal{D}_\mathrm{u} \leftarrow \mathcal{D}_\mathrm{u} \setminus \mathcal{D}_\mathrm{q}$. The model $f_{\bm{\theta}}$ is then retrained iteratively until a certain metric reaches a target value or the unlabeled data pool $\mathcal{D}_\mathrm{u}$ is exhausted.

\subsection{Query Strategy} \label{sec:sample-strat}
We utilize a least confidence strategy~\cite{Wang2014_ActiveLabelingMethod}. The least confidence query strategy is designed to acquire samples with the smallest probability among the maximum activations. Given an unlabeled data pool $\mathcal{D}_\mathrm{u}$, trained model $f_{\bm{\theta}}$, and the query size $Q$, the strategy $\alpha^\mathrm{LC}$ is defined as follows:
\begin{enumerate}
    \item Compute the posterior probability $p_\theta(y \mid \bm{x})$ for all $x \in \mathcal{D}_\mathrm{u}$ based on (\ref{eq:softmax}).
    \item Obtain the corresponding confidence $\max_{y} (p_{\bm{\theta}}(y|\bm{x})).$
    \item Sort the samples in $\mathcal{D}_\mathrm{u}$ in ascending order with respect to the confidence.
    \item The strategy $\alpha^\mathrm{LC}$ returns the top $Q$ samples.
\end{enumerate}
Based on this strategy, a set of samples is selected from the unlabeled data pool to query the oracle for annotation.

\subsection{Joint Learning of a Classifier and an Energy-based Model}
Among various methods proposed for joint learning of a classifier and an EBM~\cite{Grathwohl2019_ClassifierEnergyModel,Yang2021_JemPlusPlus,Yang2023_sadaJEM,Hayashi2023_gaussianCoupled}, we construct our model $f_{\bm{\theta}}$ with reference to the structure of JEM~\cite{Grathwohl2019_ClassifierEnergyModel}. The model consists of a single neural network with a multi-head output for classification and EBM. Given an input $\bm{x}\in\mathbb{R}^D$, the model first outputs a real-valued $K$-dimensional vector, i.e., $f_{\bm{\theta}}:\mathbb{R}^D\rightarrow\mathbb{R}^K$. The vector is then converted into the class posterior probability in the classification head and the probability density of input data in the EBM head. In the classification head, the posterior probability of class $y \in \{1, \ldots, K\}$ is calculated through the standard softmax transfer function, as defined below:
\begin{equation} \label{eq:softmax}
    p_{\bm{\theta}}(y\mid \bm{x})=\frac{\exp(f_{\bm{\theta}}(\bm{x})[y])}{\sum_{y'=1}^K\exp(f_{\bm{\theta}}(\bm{x})[y'])},
\end{equation}
where $f_{\bm{\theta}}(\bm{x})[y]$ indicates the logit corresponding to $y$-th class. In the EBM head, the probability density $p(\bm{x})$ is computed as follows:
\begin{equation} \label{eq:energy}
    p_{\bm{\theta}}(\bm{x})=\frac{\sum_{y=1}^K\exp(f_{\bm{\theta}}(\bm{x})[y])}{Z(\bm{\theta})},
\end{equation}
where $Z(\bm{\theta})=\int_{\bm{x}}\sum_{y=1}^K\exp(f_{\bm{\theta}}(\bm{x})[y])\mathrm{d}\bm{x}$ is the normalizing constant, otherwise known as the partition function.\par

In the training of the hybrid model, we minimize the following loss function over the union of labeled and unlabeled datasets $\{(\bm{x}_n, y_n)\}_{i=1}^{M}\cup\{\bm{x}_i\}_{i=M+1}^{M+N}$.
\begin{equation}
\label{eq:loss_function}
\mathcal{L} \!=\! -\sum_{i=1}^M\log p(y_i \mid \bm{x}_i) -\sum_{i=1}^{M+N}\log p(\bm{x}_{i}),
\end{equation}
where the first term on the right-hand side corresponds to cross-entropy and is optimized via conventional stochastic gradient descent. The second term is the negative log-likelihood for the EBM optimization whose gradient can be computed using stochastic gradient Langevin dynamics (SGLD). 

As a distinction from the original JEM training algorithm, we incorporate techniques from recent EBM studies to stabilize and accelerate training. First, we adopt the informative initialization~\cite{Yang2021_JemPlusPlus}, which uses samples from a Gaussian mixture distribution estimated from the training dataset instead of random noise samples for initializing the SGLD chain. Second, we employ Proximal-YOPO-SGLD~\cite{Yang2021_JemPlusPlus}, which freezes the gradient with respect to the second and subsequent layers during the sample updates. Third, we exclude data augmentation from the maximum likelihood estimation pipeline to alleviate the adverse effects of data augmentation on image generation quality~\cite{Yang2023_sadaJEM}.

\section{Experiments}
\subsection{Experimental conditions}
To verify the validity of CALICO, we conducted experiments using medical image datasets. We used five benchmark medical imaging datasets found in the MedMNIST collection \cite{yang2023_medmnist}, namely Blood, Derma, OrganS, OrganC, and Pneumonia. These medical imaging datasets consist of preprocessed 28$\times$28 two-dimensional images, accompanied by their corresponding class labels. The classification tasks within these datasets range from binary to multi-class, serving as benchmarks for foundational models in the medical imaging domain.

% The model utilized a Wide-ResNet architecture \cite{zagoruyko2016_wrn}, with hyperparameters closely following the setup of \cite{Yang2021_JemPlusPlus}. We introduced a minor modification to the Wide-ResNet backbone, specifically involving the replacement of the ReLU activation function with a Swish activation function \cite{ramachandran2018searching}. This change is made under the consideration that Swish might be more advantageous in capturing underlying patterns in the data, particularly in the presence of small negative values.

The model closely followed the setup of \cite{Yang2021_JemPlusPlus}. However, we changed the ReLU activation function used by the Wide-ResNet architecture to a Swish activation function for the added stability observed by \cite{du2019implicit}. As the computational training time of JEM++ is also relatively long, we limited the number of queries for all datasets (more details in Appendix).

We evaluated the results based on classification accuracy and calibration errors. Miscalibration can be condensed into a convenient scalar metric, often quantified as the expected calibration error (ECE) \cite{Naeini2015_WellCalibratedProbabilities}. This metric discretizes probability intervals into a fixed number of bins and the calibration error is calculated as the difference between the fraction of correct predictions (accuracy), and the mean of the probabilities in the bin (confidence). ECE computes a weighted average of this error across bins:

\begin{equation} \label{eq:ece}\text{ECE}=\sum_{m=1}^{M_\mathrm{bin}}\frac{|\mathcal{B}_m|}{N_\mathrm{data}} \mid \mathrm{acc}(\mathcal{B}_m)-\mathrm{conf}(\mathcal{B}_m)\mid
\end{equation}
\noindent where $\mathcal{B}_m$ represents the subset of samples whose predicted confidence levels lie within the interval $I_m = (\frac{m-1}{M_\mathrm{bin}}, \frac{m}{M_\mathrm{bin}}]$, $N_\mathrm{data}$ is the total number of data points, and $\mathrm{acc}(\mathcal{B}_m)$ and $\mathrm{conf}(\mathcal{B}_m)$ denote the accuracy and confidence of $\mathcal{B}_m$, respectively. For this study, we utilize ECE as the primary metric for measuring calibration, as well as reliability diagrams for visualization.

We establish the baseline reference by maximizing only the $\log{p(y|x)}$ objective through the utilization of the entire dataset, with a softmax-based classifier (named \textbf{Baseline}), involving the evaluation of calibration error using the probability outputs from the softmax activation function and calculating the ECE. We also used a softmax-based classifier paired with an AL framework using the least confidence query strategy (named \textbf{Active}) as an additional baseline reference.

\subsection{Performance Comparison}

\begin{table}[t]
\centering
\caption{Performance Comparison of Test Accuracy ($\uparrow$) and ECE ($\downarrow$) Values}
\label{table:ece-values}
\begin{tabular}{llccc} 
\toprule
\textbf{Dataset} & \multicolumn{1}{c}{\textbf{Criterion (\%)}}        & \textbf{Baseline}     & \textbf{Active}        & \textbf{CALICO}         \\ 
\midrule
Blood            & \textbf{Best ACC~/~ECE}                            & 95.82~/~0.95          & 95.47~/~1.73           & \textbf{96.43~/~0.54}   \\
                 & \textbf{Final ACC~/~ECE}                           & -                     & 95.44~/~1.87           & \textbf{96.00 /~0.56}   \\ 
\midrule
Derma            & \textbf{\textbf{Best}~ACC~/~ECE}                   & 74.02~/~3.50          & 74.87~/~\textbf{1.82}  & \textbf{76.37~}/~1.89   \\
                 & \textbf{Final ACC~/~ECE}                           & -                     & 74.56~/~\textbf{1.82}  & \textbf{75.66 }/~1.89   \\ 
\midrule
OrganS           & \textbf{\textbf{Best}~ACC~/~ECE}                   & 78.54~/~\textbf{6.75} & 78.55~/~10.04          & \textbf{79.90~}/~7.00   \\
                 & \textbf{Final ACC~/~ECE}                           & -                     & 78.55~/~10.04          & \textbf{79.90} /~7.00   \\ 
\midrule
OrganC           & \textbf{\textbf{\textbf{\textbf{Best~}}}ACC~/~ECE} & 89.54~/~5.31          & 89.79~/~4.17           & \textbf{90.34~/~3.03}   \\
                 & \textbf{Final ACC~/~ECE}                           & -                     & 88.58~/~5.76           & \textbf{90.26 /~3.03}   \\ 
\midrule
Pneumonia        & \textbf{Best ACC~/~ECE}                            & 87.66~/~12.33         & \textbf{89.26~/~9.73}  & \textbf{89.26}~/~9.82   \\
                 & \textbf{Final ACC~/~ECE}                           & -                     & \textbf{87.66}~/~12.33 & 86.22~/~\textbf{11.88}  \\
\bottomrule
\end{tabular}
\end{table}

% \begin{table}[t]
% \centering
% \caption{Performance Comparison of Test Accuracy ($\uparrow$) and ECE ($\downarrow$) Values}
% \label{table:ece-values}
% \begin{tabular}{lcccc}
% \toprule
% \textbf{Dataset} & \textbf{Criterion (\%)} & \textbf{Baseline / Active / CALICO} \
% \midrule
% Blood & \textbf{Max. ACC~/ECE} & 95.82/0.95 & 95.47/1.73 & \textbf{96.43/0.54} \
% & \textbf{Final ACC/ECE} & -/1.87 & 95.44/1.82 & \textbf{-/0.56} \
% \midrule
% Derma & \textbf{Max. ACC/ECE} & 74.02/3.50 & 74.87/1.82 & \textbf{76.37/1.89} \
% & \textbf{Final ACC/ECE} & -/1.82 & 74.56/1.82 & \textbf{-/1.89} \
% \midrule
% OrganS & \textbf{Max. ACC/ECE} & 78.54/6.75 & 78.55/10.04 & \textbf{79.90/7.00} \
% & \textbf{Final ACC/ECE} & -/10.04 & 78.55/10.04 & \textbf{-/7.00} \
% \midrule
% OrganC & \textbf{Max. ACC/ECE} & 89.54/5.31 & 89.79/4.17 & \textbf{90.34/3.03} \
% & \textbf{Final ACC/ECE} & -/5.76 & 88.58/5.76 & \textbf{-/3.03} \
% \midrule
% Pneumonia & \textbf{Max. ACC/ECE} & 87.66/12.33 & \textbf{89.26/9.73} & \textbf{89.26/9.82} \
% & \textbf{Final ACC/ECE} & -/12.33 & \textbf{87.66/12.33} & 86.22/~\textbf{11.88} \
% \bottomrule
% \end{tabular}
% \end{table}

Table \ref{table:ece-values} shows that CALICO consistently outperformed the baseline accuracy across all evaluated datasets. In parallel with the accuracy improvements, CALICO also showed reduced ECEs, indicating its effectiveness in minimizing miscalibration. We observed that using a softmax-based classifier within an AL paradigm, as opposed to straightforward training methods, resulted in better calibration. However, CALICO demonstrated a more substantial increase in performance when compared to the baseline, suggesting its effectiveness in improving the classifier's performance in an AL paradigm. While CALICO generally achieved lower ECE values across most datasets, the use of a softmax-based classifier in an AL paradigm demonstrated comparable efficacy to CALICO in some cases, such as Derma and OrganS. 

\subsection{Confidence Calibration}
\begin{figure}[t]
    \centering
    \resizebox{\columnwidth}{!}{
        \input{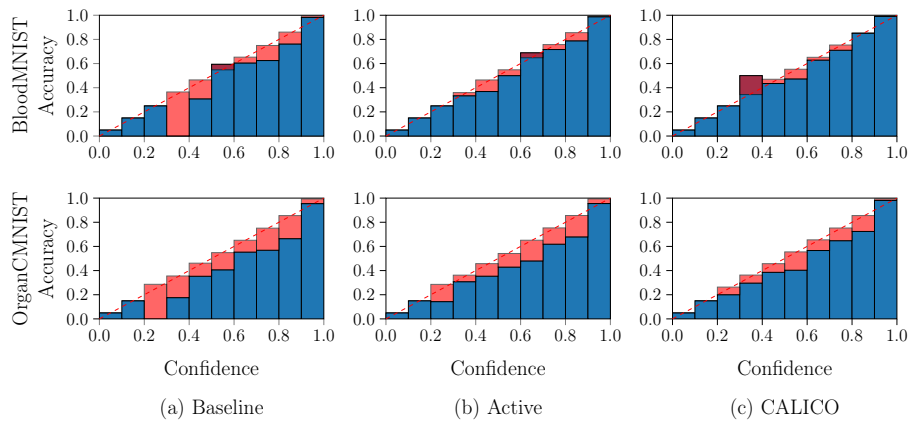}
    }
    \caption{The final reliability diagrams for the baseline, active, and CALICO. In comparison, CALICO demonstrated a substantial improvement in calibration across confidence intervals when compared to the other evaluated methods that used softmax-based classifiers.}
    \label{fig:rel-diagrams}
\end{figure}
The final reliability diagrams are depicted in Fig.~\ref{fig:rel-diagrams}. A perfectly calibrated model would align with the red-diagonal dashed line, effectively creating an identity function between the y-axis (accuracy) and the x-axis (confidence intervals). Any deviation above or below this diagonal indicates underconfidence or overconfidence, respectively, representing miscalibration. It is prominent that there is a marginal difference in the improvement of overconfident intervals between using softmax-based classifiers and CALICO, emphasizing the effectiveness of calibrated confidence outputs for iterative training in an AL paradigm.
\begin{figure}[!t]
    \centering
    \resizebox{\columnwidth}{!}{
        \input{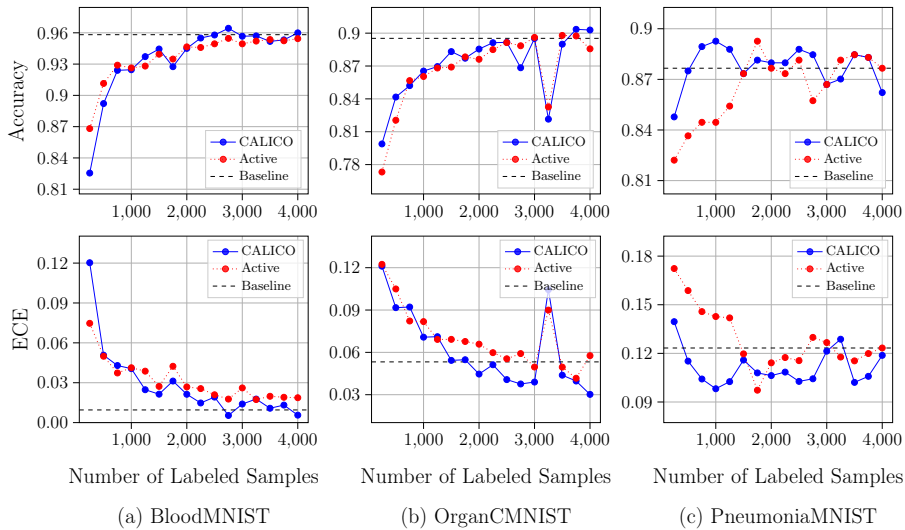}
    }
    \caption{The test accuracy (top) and ECE (bottom) values plotted against the number of labeled samples per AL iteration. It can be observed that a lower or comparable ECE value can be achieved with a lesser number of labeled samples.}
    \label{fig:calibration-curves}
\end{figure}

\subsection{Learning Curve Analysis}
Learning curves play a crucial role in assessing how well a model performs in an AL paradigm, particularly by observing a specific metric with respect to the number of labeled samples. In this study, the ECE is of certain importance. Similar to the commonly used accuracy, the goal is to determine whether miscalibration can be minimized with significantly fewer labeled samples through the utilization of JEMs. As illustrated in Fig. \ref{fig:calibration-curves}, the overall calibration trend using CALICO is observed to be lower than that of the softmax-based classifiers. Additionally, it is noteworthy that a lower or comparable ECE can be achieved with fewer samples in comparison to the baseline reference.

\subsection{Performance Comparison of Test Accuracy and ECE Values with an Equal Class Distribution}
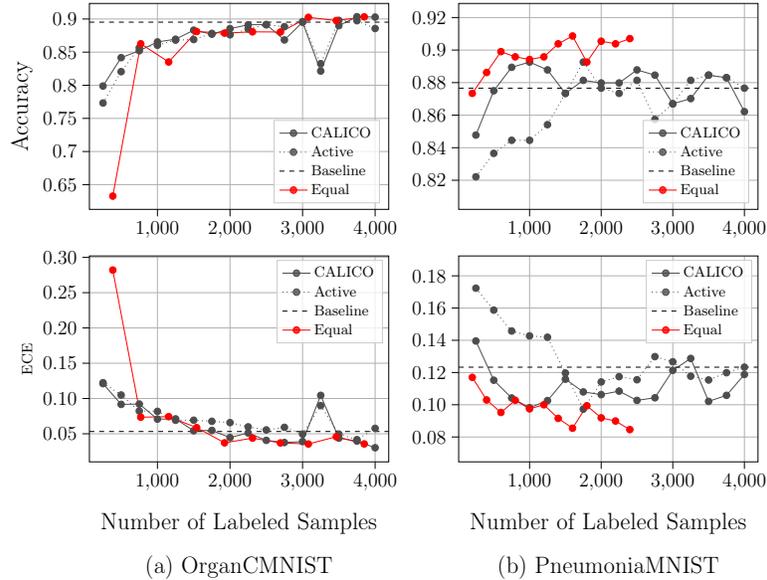
\begin{figure}[t]
    \centering
    \resizebox{0.85\columnwidth}{!}{
        % This file was created by tikzplotlib v0.9.8.
\begin{tikzpicture}

\begin{groupplot}[group style={group size=2 by 2, vertical sep=1cm, horizontal sep=1.5cm}, height=6cm, width=8cm]
\nextgroupplot[
legend cell align={left},
legend style={
  fill opacity=0.8,
  draw opacity=1,
  text opacity=1,
  at={(0.97,0.03)},
  anchor=south east,
  draw=white!80!black
},
tick align=outside,
tick pos=left,
x grid style={white!69.0196078431373!black},
xmajorgrids,
xmin=62.5, xmax=4187.5,
xtick style={color=black},
y grid style={white!69.0196078431373!black},
ylabel={Accuracy},
ymajorgrids,
ymin=0.612680237293243, ymax=0.923483331203461,
ytick style={color=black},
ytick distance=0.05,
xlabel style={font=\Large},
ylabel style={font=\Large},
yticklabel style={font=\large},
xticklabel style={font=\large},
]
\addplot [semithick, black!70, mark=*, mark size=2, mark options={solid}]
table {%
250 0.798863112926483
500 0.841678738594055
750 0.852080285549164
1000 0.865384638309479
1250 0.869496881961823
1500 0.88316398859024
1750 0.877116620540619
2000 0.885462045669556
2250 0.891267538070679
2500 0.891630411148071
2750 0.86840832233429
3000 0.895137906074524
3250 0.821480393409729
3500 0.889937102794647
3750 0.903483331203461
4000 0.90287858247757
};
\addlegendentry{CALICO}
\addplot [semithick, black!70, dotted, mark=*, mark size=2, mark options={solid}]
table {%
250 0.773222088813782
500 0.82051283121109
750 0.856676340103149
1000 0.860425710678101
1250 0.868166446685791
1500 0.869013071060181
1750 0.878326058387756
2000 0.876148998737335
2250 0.884978234767914
2500 0.891388475894928
2750 0.888485729694366
3000 0.896105468273163
3250 0.832607626914978
3500 0.897919714450836
3750 0.897435903549194
4000 0.885703921318054
};
\addlegendentry{Active}
\addplot [semithick, black, dashed]
table {%
62.5 0.895258843898773
4187.5 0.895258843898773
};
\addlegendentry{Baseline}
\addplot [semithick, red, mark=*, mark size=2, mark options={solid}]
table {%
385 0.632680237293243
770 0.862602829933167
1155 0.835147559642792
1540 0.881107866764069
1925 0.878688931465149
2310 0.880624115467072
2695 0.880382180213928
3080 0.902515709400177
3465 0.897677779197693
3850 0.903241395950317
};
\addlegendentry{Equal}

\nextgroupplot[
legend cell align={left},
legend style={
  fill opacity=0.8,
  draw opacity=1,
  text opacity=1,
  at={(0.97,0.03)},
  anchor=south east,
  draw=white!80!black
},
tick align=outside,
tick pos=left,
x grid style={white!69.0196078431373!black},
xmajorgrids,
xmin=10, xmax=4190,
xtick style={color=black},
y grid style={white!69.0196078431373!black},
ymajorgrids,
ymin=0.802115361690521, ymax=0.928653855323792,
ytick style={color=black},
ytick distance=0.02,
xlabel style={font=\Large},
ylabel style={font=\Large},
yticklabel style={font=\large},
xticklabel style={font=\large},
]
\addplot [semithick, black!70, mark=*, mark size=2, mark options={solid}]
table {%
250 0.847756385803223
500 0.875
750 0.889423072338104
1000 0.892628192901611
1250 0.887820541858673
1500 0.873397409915924
1750 0.881410241127014
2000 0.879807710647583
2250 0.879807710647583
2500 0.887820541858673
2750 0.884615361690521
3000 0.86698716878891
3250 0.870192289352417
3500 0.884615361690521
3750 0.88301283121109
4000 0.862179458141327
};
\addlegendentry{CALICO}
\addplot [semithick, black!70, dotted, mark=*, mark size=2, mark options={solid}]
table {%
250 0.822115361690521
500 0.836538434028625
750 0.844551265239716
1000 0.844551265239716
1250 0.854166686534882
1500 0.873397409915924
1750 0.892628192901611
2000 0.876602590084076
2250 0.873397409915924
2500 0.881410241127014
2750 0.857371807098389
3000 0.86698716878891
3250 0.881410241127014
3500 0.884615361690521
3750 0.88301283121109
4000 0.876602590084076
};
\addlegendentry{Active}
\addplot [semithick, black, dashed]
table {%
10.0000000000002 0.876602590084076
4190 0.876602590084076
};
\addlegendentry{Baseline}
\addplot [semithick, red, mark=*, mark size=2, mark options={solid}]
table {%
200 0.873397409915924
400 0.886217951774597
600 0.899038434028625
800 0.895833313465118
1000 0.894230782985687
1200 0.895833313465118
1400 0.903846144676208
1600 0.908653855323792
1800 0.892628192901611
2000 0.905448734760284
2200 0.903846144676208
2400 0.907051265239716
};
\addlegendentry{Equal}

\nextgroupplot[
legend cell align={left},
legend style={fill opacity=0.8, draw opacity=1, text opacity=1, draw=white!80!black},
tick align=outside,
tick pos=left,
x grid style={white!69.0196078431373!black},
xlabel style={font=\Large\linespread{1.5}\selectfont, yshift=-0.25cm,align=center},
xlabel={Number of Labeled Samples\\(a) OrganCMNIST},
xmajorgrids,
xmin=62.5, xmax=4187.5,
xtick style={color=black},
y grid style={white!69.0196078431373!black},
ylabel={ECE},
ymajorgrids,
ymin=0.0102237819882049, ymax=0.302041555394008,
ytick style={color=black},
yticklabel style={
    /pgf/number format/fixed,
    /pgf/number format/fixed zerofill,
    /pgf/number format/precision=2,
    font=\large
},
ytick distance=0.05,
xticklabel style={font=\large},
]
\addplot [semithick, black!70, mark=*, mark size=2, mark options={solid}]
table {%
250 0.121067874820251
500 0.0916890278728935
750 0.0921325724270964
1000 0.0707641166401013
1250 0.0711196301821268
1500 0.0543115359785537
1750 0.0547812935370405
2000 0.044619462684871
2250 0.0512252902647234
2500 0.0406935078654678
2750 0.0376108539136181
3000 0.0389569385917105
3250 0.104448011659543
3500 0.0438349614875957
3750 0.0396911380573724
4000 0.0302237819882049
};
\addlegendentry{CALICO}
\addplot [semithick, black!70, dotted, mark=*, mark size=2, mark options={solid}]
table {%
250 0.122377477519585
500 0.104952840497659
750 0.0822218456433974
1000 0.0817967112017069
1250 0.0691739151521933
1500 0.0692183809669397
1750 0.0677123884893467
2000 0.0658020863159013
2250 0.0598023568388627
2500 0.0554599688592513
2750 0.0591421059878424
3000 0.0496050337456566
3250 0.0899571895772383
3500 0.0495574903815451
3750 0.041664597268807
4000 0.0576723532460136
};
\addlegendentry{Active}
\addplot [semithick, black, dashed]
table {%
62.5 0.0532503908938928
4187.5 0.0532503908938928
};
\addlegendentry{Baseline}
\addplot [semithick, red, mark=*, mark size=2, mark options={solid}]
table {%
385 0.282041555394008
770 0.0734876733242628
1155 0.0742817021668476
1540 0.0587314237314661
1925 0.0372055137325411
2310 0.0438769018835328
2695 0.037072366928068
3080 0.0355541767939081
3465 0.045836567750762
3850 0.0356572221858264
};
\addlegendentry{Equal}

\nextgroupplot[
legend cell align={left},
legend style={fill opacity=0.8, draw opacity=1, text opacity=1, draw=white!80!black},
tick align=outside,
tick pos=left,
x grid style={white!69.0196078431373!black},
xlabel style={font=\Large\linespread{1.5}\selectfont, yshift=-0.25cm,align=center},
xlabel={Number of Labeled Samples\\(b) PneumoniaMNIST},
xmajorgrids,
xmin=10, xmax=4190,
xtick style={color=black},
y grid style={white!69.0196078431373!black},
ymajorgrids,
ymin=0.0645714916461163, ymax=0.192336402890645,
ytick style={color=black},
yticklabel style={
    /pgf/number format/fixed,
    /pgf/number format/fixed zerofill,
    /pgf/number format/precision=2,
    font=\large
},
ytick distance=0.02,
ylabel style={font=\Large},
xticklabel style={font=\large},
]
\addplot [semithick, black!70, mark=*, mark size=2, mark options={solid}]
table {%
250 0.139557288651129
500 0.115225150172066
750 0.104184235603567
1000 0.0981726749066436
1250 0.102549841996008
1500 0.115866636745469
1750 0.107952312450091
2000 0.106295912934724
2250 0.108450589826927
2500 0.102709747519419
2750 0.104318561625848
3000 0.121360382470765
3250 0.128732975391545
3500 0.102095171802655
3750 0.105833198279237
4000 0.118779369196184
};
\addlegendentry{CALICO}
\addplot [semithick, black!70, dotted, mark=*, mark size=2, mark options={solid}]
table {%
250 0.172336402890645
500 0.158731931223841
750 0.145763470785543
1000 0.142727371509886
1250 0.141893528956183
1500 0.119684943778275
1750 0.0972822668784798
2000 0.114158021110602
2250 0.117430054628022
2500 0.115513719889924
2750 0.129867247445954
3000 0.126669803625728
3250 0.117719268400178
3500 0.115370588632358
3750 0.119889003774179
4000 0.123319206300934
};
\addlegendentry{Active}
\addplot [semithick, black, dashed]
table {%
10 0.12331331743378
4190 0.12331331743378
};
\addlegendentry{Baseline}
\addplot [semithick, red, mark=*, mark size=2, mark options={solid}]
table {%
200 0.116999322753912
400 0.102972955284945
600 0.0952168258745699
800 0.102781007307454
1000 0.0974503194565766
1200 0.0999945157750145
1400 0.0915448673575464
1600 0.0854422866141242
1800 0.0993648804950933
2000 0.0918807663030421
2200 0.0899024620364486
2400 0.0845714916461163
};
\addlegendentry{Equal}
\end{groupplot}

% \draw ({$(current bounding box.south west)!0.5!(current bounding box.south east)$}|-{$(current bounding box.south west)!0.5!(current bounding box.north west)$}) node[
%   scale=1.75,
%   text=black,
%   rotate=0.0
% ]{(a) OrganCMNIST};
% \draw ({$(current bounding box.south west)!0.5!(current bounding box.south east)$}|-{$(current bounding box.south west)!-0.05!(current bounding box.north west)$}) node[
%   scale=1.75,
%   text=black,
%   rotate=0.0
% ]{(b) PneumoniaMNIST};
\end{tikzpicture}
    }
    \caption{The test accuracy (top) and ECE (bottom) values of CALICO with an equal class distribution. Note that the \textit{Equal} method was designed to strictly enforce an equal distribution by setting a limit based on the dataset's class label with the fewest samples. Additionally, the number of labels per class varied per dataset to allow for an adequate number of iterations to properly analyze the learning curve.}
    \label{fig:ablation}
\end{figure}

% \begin{table}[!ht]
% \centering
% \caption{Min. and Final Test Accuracy and ECE Values in Comparison to an Equal Class Distribution}
% \label{table:ablation-values}
% \begin{tabular}{lcccc} 
% \toprule
% \textbf{Dataset} & \textbf{Criterion (\%)} & \textbf{Baseline} & \textbf{CALICO} & \multicolumn{1}{l}{\textbf{Equal}} \\ 
% \hline
% Blood & \textbf{Max. ACC $\uparrow$} & 95.82 & \textbf{96.43} & 96.26 \\
%  & \textbf{Final ACC $\uparrow$} & - & 96.00 & \textbf{96.05} \\ 
% \cline{2-5}
%  & \textbf{Min. ECE $\downarrow$} & 0.95 & \textbf{0.54} & 0.89 \\
%  & \textbf{Final ECE $\downarrow$} & - & \textbf{0.56} & 0.89 \\ 
% \midrule
% OrganS & \textbf{Max. ACC $\uparrow$} & 78.54 & \textbf{79.90} & 78.98 \\
%  & \textbf{Final ACC $\uparrow$} & - & \textbf{79.90} & 78.07 \\ 
% \cline{2-5}
%  & \textbf{Min. ECE $\downarrow$} & \textbf{6.75} & 7.00 & 10.49 \\
%  & \textbf{Final ECE $\downarrow$} & - & 7.00 & 10.49 \\ 
% \midrule
% OrganC & \textbf{Max. ACC $\uparrow$} & 89.54 & \textbf{90.34} & 90.25 \\
%  & \textbf{Final ACC $\uparrow$} & - & 90.26 & \textbf{90.32} \\ 
% \cline{2-5}
%  & \textbf{Min. ECE $\downarrow$} & 5.31 & \textbf{3.03} & 3.56 \\
%  & \textbf{Final ECE $\downarrow$} & - & \textbf{3.03} & 3.57 \\ 
% \midrule
% Pneumonia & \textbf{Max. ACC $\uparrow$} & 87.66 & 89.26 & \textbf{90.87} \\
%  & \textbf{Final ACC $\uparrow$} & - & 86.22 & \textbf{90.71} \\ 
% \cline{2-5}
%  & \textbf{Min. ECE $\downarrow$} & 12.33 & 9.82 & \textbf{8.46} \\
%  & \textbf{Final ECE $\downarrow$} & - & 11.88 & \textbf{8.46} \\
% \bottomrule
% \end{tabular}
% \end{table}

\begin{table}[t]
\centering
\caption{Min. and Final Test Accuracy ($\uparrow$) and ECE ($\downarrow$) Values in Comparison to an Equal Class Distribution}
\label{table:ablation-values}
\begin{tabular}{llccc} 
\toprule
\textbf{Dataset} & \multicolumn{1}{c}{\textbf{Criterion (\%)}} & \textbf{Baseline} & \textbf{CALICO}       & \textbf{\textbf{Equal}}  \\ 
\midrule
Blood            & \textbf{Best ACC~/~ECE}                     & 95.82~/~0.95      & \textbf{96.43~/~0.54} & 96.25 / 0.89             \\
                 & \textbf{Final ACC~/~ECE}                    & -                 & 96.00\textbf{ /~0.56} & \textbf{96.05} / 0.89    \\ 
\midrule
OrganS           & \textbf{\textbf{Best}~ACC~/~ECE}            & 78.54~/~6.75      & \textbf{79.90~/~7.00} & 78.98 / 10.49            \\
                 & \textbf{Final ACC~/~ECE}                    & -                 & \textbf{79.90~/~7.00} & 78.07 / 10.49            \\ 
\midrule
OrganC           & \textbf{\textbf{Best}~ACC~/~ECE}            & 89.54~/~5.31      & \textbf{90.34~/~3.03} & 90.25 / 3.56             \\
                 & \textbf{Final ACC~/~ECE}                    & -                 & 90.26 /~\textbf{3.03} & \textbf{90.32 }/ 3.57    \\ 
\midrule
Pneumonia        & \textbf{\textbf{Best}~ACC~/~ECE}            & 87.66~/~12.33     & 89.26~/~9.82          & \textbf{90.87 / 8.46}    \\
                 & \textbf{Final ACC~/~ECE}                    & -                 & 86.22~/~11.88         & \textbf{90.71 / 8.46}    \\
\bottomrule
\end{tabular}
\end{table}

We explored the impact of class distribution balancing on CALICO's performance. While many confidence calibration studies emphasize calibration error within balanced class distributions, achieving such balance is not guaranteed in AL due to sampling methods' greedy nature. Calibration learning curves (Fig. \ref{fig:calibration-curves}) exhibited instability, possibly due to overconfidence from uncertainty-based sampling. Additionally, inherent dataset class imbalances can lead to querying uninformative samples, creating a mode collapse problem and miscalibration. To simulate literature setups, CALICO's performance was evaluated with an equal class distribution (named \textbf{Equal}), enforcing limits based on the dataset's least represented class. Varying labels per class allowed for sufficient iterations to analyze learning curves. Experimental setup details are provided in Table \ref{table:ablation-limits} of Appendix.

% \begin{table}[!hb]
% \centering
% \caption{Comparison of Min. and Final Test Accuracy and ECE Values with Equal Class Distribution}
% \label{table:ablation-values}
% \begin{tabular}{lccccc}
% \toprule
% \textbf{Dataset} & \multicolumn{2}{c}{\textbf{Accuracy (\%)}} & \multicolumn{2}{c}{\textbf{ECE}} \\
% \cmidrule(lr){2-3} \cmidrule(lr){4-5}
% & \textbf{Max.} & \textbf{Final} & \textbf{Min.} & \textbf{Final} \\
% \midrule
% Blood & 95.82 / 96.43 & - / 96.00 & 0.95 / 0.54 & - / 0.56 \\
% OrganS & 78.54 / 79.90 & - / 79.90 & 6.75 / 7.00 & - / 7.00 \\
% OrganC & 89.54 / 90.34 & - / 90.26 & 5.31 / 3.03 & - / 3.03 \\
% Pneumonia & 87.66 / 89.26 & - / 86.22 & 12.33 / 9.82 & - / 11.88 \\
% \bottomrule
% \end{tabular}
% \end{table}

We observed instances where having an equal class distribution yielded better calibration or more stable learning curves across the evaluated datasets, such as the results on the PneumoniaMNIST dataset. However, Table \ref{table:ablation-values} also highlights instances where equal class distribution did not result in better calibration compared to the original CALICO. One possible explanation for this disparity is the nature of the datasets; PneumoniaMNIST is an imbalanced binary dataset, while others are relatively balanced multi-class datasets, and balancing the class ratio in sample selection facilitated effective learning of information from the minority class. These findings imply the potential of class distribution balancing to enhance CALICO's performance on imbalanced datasets.
% It is crucial to note that this ablation study compares CALICO to a best-case scenario, and does not reflect the true performance in real-world applications. Therefore, combining the use of JEMs in an AL paradigm could prove advantageous in minimizing overall miscalibration when dealing with limited labeled data. 

\section{Conclusion}
We proposed an AL method called CALICO, which aimed to use the calibrated confidence outputs as the input for a query strategy in an AL paradigm. CALICO incorporates the joint training of a classifier and en EBM, allowing self-calibration of confidence used for sample selection in AL. Experimental results demonstrated that CALICO outperformed the baseline accuracy and achieved a lower ECE with less labeled data, compared to a softmax-based classifier. 
% Moreover, we address the issue of limited labeled data, incorporating the use of JEMs for self-calibration and eliminating the need for a separate labeled dataset, as commonly required by post-hoc calibration methods. This makes the utilization of available data more efficient. 

% One of the immediate limitations of this study lies in using JEMs. Each iteration required a substantial amount of time, limiting the number of experiments that could be conducted. The stability of JEMs is also dependent on the chosen hyperparameters. Additionally, 
One of the limitations of this study is scalability because training an EBM on high-resolution images requires considerable hyperparameter tuning, and CALICO was only evaluated on small-resolution images. Future research could explore the applications of CALICO with other AL methods, harnessing the power of Bayesian approaches for better uncertainty quantification. Additionally, further evaluation of larger datasets in other domains can broaden the scope and applicability of this research domain.

\section*{Appendix}
% \section{Experimental Details} \label{app:exp-details}
% \subsection*{Experimental Setup}
\paragraph{Experimental Setup} To ensure consistency across all experiments, the computational runtime constraints necessitated limiting each dataset to 4000 labeled samples, with a query size of 250 for each iteration, resulting in a total of 16 iterations. In the ablation study that focused on an equal class distribution, the experimental setup was detailed in Table \ref{table:ablation-limits}. The number of labeled samples per class for each iteration was determined by the class with the fewest samples to create enough iterations for analysis. The only exception to the ablation study was the DermaMNIST dataset, where the class with the lowest number of samples was 89. It was decided not to include this dataset in the ablation study.

% \begin{table}
% \centering
% \caption{Experimental Setup for Active Learning}
% \label{table:exp-limits}
% \begin{tabular}{lccc} 
% \toprule
% \textbf{Dataset} & \textbf{Original Train Set Size} & \textbf{Limit} & \textbf{Query Size}  \\ 
% \midrule
% Blood      & 11,959 & 4,000 & 250 \\
% OrganS     & 13,940 & 4,000 & 250 \\
% OrganC     & 12,975 & 4,000 & 250 \\
% Derma      & 7,007  & 4,000 & 250 \\
% Pneumonia  & 4,708  & 4,000 & 250 \\
% \bottomrule
% \end{tabular}
% \end{table}

\begin{table}[t]
\centering
\caption{Experimental Setup for Equal Class Distribution}
\label{table:ablation-limits}
\begin{tabular}{lccc} 
\toprule
\textbf{Dataset} & \textbf{Lowest Count (Class)} & \textbf{Limit} & \textbf{Labels per Class}  \\ 
\midrule
Blood      & 849 (lymphocyte)        & 4,000 & 50  \\
OrganS     & 614 (femur-right)       & 3,850 & 35  \\
OrganC     & 600 (heart)             & 3,850 & 35  \\
Pneumonia  & 1,214 (normal)          & 2,400 & 100 \\
\bottomrule
\end{tabular}
\end{table}

% \begin{table}[!ht]
% \centering
% \caption{Experimental Setup for Active Learning and Equal Class Distribution}
% \label{table:combined-limits}
% \begin{tabular}{lcccccc} 
% \toprule
% \multirow{2}{*}{\textbf{Dataset}} & \multicolumn{3}{c}{\textbf{Active Learning}} & \multicolumn{3}{c}{\textbf{Equal Class Distribution}} \\ 
% \cmidrule(lr){2-4} \cmidrule(lr){5-7}
% & \textbf{Original Train Set Size} & \textbf{Limit} & \textbf{Query Size} & \textbf{Lowest Count (Class)} & \textbf{Limit} & \textbf{Labels per Class} \\ 
% \midrule
% Blood      & 11,959 & 4,000 & 250 & 849 (lymphocyte) & 4,000 & 50  \\
% OrganS     & 13,940 & 4,000 & 250 & 614 (femur-right) & 3,850 & 35  \\
% OrganC     & 12,975 & 4,000 & 250 & 600 (heart) & 3,850 & 35  \\
% Derma      & 7,007  & 4,000 & 250 & - & - & - \\
% Pneumonia  & 4,708  & 4,000 & 250 & 1,214 (normal) & 2,400 & 100 \\
% \bottomrule
% \end{tabular}
% \end{table}

% \subsection*{Hyperparameter Settings} \label{app-sub:hparams}
\paragraph{Hyperparameter Settings} All datasets, except for PneumoniaMNIST, adapted the default hyperparameters from the original literature of JEM++ \cite{Yang2021_JemPlusPlus}. This included using an SGD optimizer with a learning rate of 0.1. However, for PneumoniaMNIST, an Adam optimizer with a learning rate of 0.0001 was utilized, as it exhibited a more stable calibration performance.

\begin{credits}
\subsubsection{\ackname} This work was supported by JSPS KAKENHI Grant Number JP24K03010 and the World Premier International Research Center Initiative (WPI), MEXT, Japan.

% \subsubsection{\discintname}
% It is now necessary to declare any competing interests or to specifically
% state that the authors have no competing interests. Please place the
% statement with a bold run-in heading in small font size beneath the
% (optional) acknowledgments\footnote{If EquinOCS, our proceedings submission
% system, is used, then the disclaimer can be provided directly in the system.},
% for example: The authors have no competing interests to declare that are
% relevant to the content of this article. Or: Author A has received research
% grants from Company W. Author B has received a speaker honorarium from
% Company X and owns stock in Company Y. Author C is a member of committee Z.
\end{credits}
\bibliographystyle{splncs04}
\bibliography{cites}

\begin{thebibliography}{10}
\providecommand{\url}[1]{\texttt{#1}}
\providecommand{\urlprefix}{URL }
\providecommand{\doi}[1]{https://doi.org/#1}

\bibitem{abdar2021_reviewUQ}
Abdar, M., Pourpanah, F., Hussain, S., Rezazadegan, D., Liu, L., Ghavamzadeh, M., Fieguth, P., Cao, X., Khosravi, A., Acharya, U.R., et~al.: A review of uncertainty quantification in deep learning: Techniques, applications and challenges. Information fusion  \textbf{76},  243--297 (2021)

\bibitem{bakhtin2021residual}
Bakhtin, A., Deng, Y., Gross, S., Ott, M., Ranzato, M., Szlam, A.: Residual energy-based models for text. Journal of Machine Learning Research  \textbf{22}(40),  1--41 (2021)

\bibitem{Budd2021_MedicalImageAL}
Budd, S., Robinson, E.C., Kainz, B.: A survey on active learning and human-in-the-loop deep learning for medical image analysis. Medical Image Analysis  \textbf{71},  102062 (2021)

\bibitem{deng2020residual}
Deng, Y., Bakhtin, A., Ott, M., Szlam, A., Ranzato, M.: Residual energy-based models for text generation. In: Proceedings of the International Conference on Learning Representations (2020)

\bibitem{du2019implicit}
Du, Y., Mordatch, I.: Implicit generation and modeling with energy based models. In: Proceedings of the Annual Conference on Neural Information Processing Systems. vol.~32 (2019)

\bibitem{Fu2012_InstanceSelection}
Fu, Y., Zhu, X., Li, B.: A survey on instance selection for active learning. Knowledge and Information Systems  \textbf{35},  249--283 (2013)

\bibitem{Gal2015_BayesCNN}
Gal, Y., Ghahramani, Z.: {Bayesian} convolutional neural networks with {Bernoulli} approximate variational inference. arXiv preprint arXiv:1506.02158  (2015)

\bibitem{Gal2017_BayesianActiveLearning}
Gal, Y., Islam, R., Ghahramani, Z.: Deep {Bayesian} active learning with image data. In: {Proceedings of the International Conference on Machine Learning}. pp. 1183--1192. PMLR (2017)

\bibitem{Grathwohl2019_ClassifierEnergyModel}
Grathwohl, W., Wang, K.C., Jacobsen, J.H., Duvenaud, D., Norouzi, M., Swersky, K.: Your classifier is secretly an energy based model and you should treat it like one. In: {Proceedings of the International Conference on Learning Representations} (2020)

\bibitem{Guo2017_calibrationModern}
Guo, C., Pleiss, G., Sun, Y., Weinberger, K.Q.: On calibration of modern neural networks. In: {In Proceedings of the International Conference on Machine Learning}. pp. 1321--1330 (2017)

\bibitem{Hayashi2023_gaussianCoupled}
Hayashi, H.: A hybrid of generative and discriminative models based on the {Gaussian}-coupled softmax layer. IEEE Transactions on Neural Networks and Learning Systems (Early Access)  (2024)

\bibitem{hein2022_comparisonUQ}
Hein, A., R{\"o}hrl, S., Grobel, T., Lengl, M., Hafez, N., Knopp, M., Klenk, C., Heim, D., Hayden, O., Diepold, K.: A comparison of uncertainty quantification methods for active learning in image classification. In: {Proceedings of the International Joint Conference on Neural Networks}. pp.~1--8 (2022)

\bibitem{Kumar2020_ALQueryStrategies}
Kumar, P., Gupta, A.: Active learning query strategies for classification, regression, and clustering: a survey. Journal of Computer Science and Technology  \textbf{35},  913--945 (2020)

\bibitem{LeCun2006_EnergyBasedLearning}
LeCun, Y., Chopra, S., Hadsell, R., Ranzato, M., Huang, F.: A tutorial on energy-based learning. Predicting structured data  \textbf{1} (2006)

\bibitem{Liu2022_ssmJEM}
Liu, X., Staudt, D., Lin, C.T., Zach, C.: Effortless training of joint energy-based models with sliced score matching. In: {Proceedings of the International Conference on Pattern Recognition}. pp. 2643--2649 (2022)

\bibitem{minderer2021_revisitCalibration}
Minderer, M., Djolonga, J., Romijnders, R., Hubis, F., Zhai, X., Houlsby, N., Tran, D., Lucic, M.: Revisiting the calibration of modern neural networks. In: Proceedings of the Advances in Neural Information Processing Systems. vol.~34, pp. 15682--15694 (2021)

\bibitem{Naeini2015_WellCalibratedProbabilities}
Naeini, M.P., Cooper, G., Hauskrecht, M.: Obtaining well calibrated probabilities using {Bayesian} binning. In: {Proceedings of the AAAI Conference on Artificial Intelligence}. vol.~29 (2015)

\bibitem{Ren2021_DeepALSurvey}
Ren, P., Xiao, Y., Chang, X., Huang, P.Y., Li, Z., Gupta, B.B., Chen, X., Wang, X.: A survey of deep active learning. ACM Computing Surveys  \textbf{54},  1--40 (2021)

\bibitem{salehinejad2021edropout}
Salehinejad, H., Valaee, S.: {EDropout}: Energy-based dropout and pruning of deep neural networks. IEEE Transactions on Neural Networks and Learning Systems  \textbf{33}(10),  5279--5292 (2022)

\bibitem{Settles2009_ActiveLearningSurvey}
Settles, B.: Active learning literature survey. Computer Sciences Technical Report~1648, University of Wisconsin--Madison (2009)

\bibitem{Simeoni2019_SSLActive}
Sim{\'e}oni, O., Budnik, M., Avrithis, Y., Gravier, G.: Rethinking deep active learning: Using unlabeled data at model training. In: {Proceedings of the International Conference on Pattern Recognition}. pp. 1220--1227 (2021)

\bibitem{thulasidasan2019mixup-calib}
Thulasidasan, S., Chennupati, G., Bilmes, J.A., Bhattacharya, T., Michalak, S.: On mixup training: Improved calibration and predictive uncertainty for deep neural networks. Proceedings of the Annual Conference on Neural Information Processing Systems  \textbf{32} (2019)

\bibitem{Tran2019_BayesGenerative}
Tran, T., Do, T.T., Reid, I., Carneiro, G.: {Bayesian} generative active deep learning. In: {Proceedings of the International Conference on Machine Learning}. pp. 6295--6304 (2019)

\bibitem{Wang2014_ActiveLabelingMethod}
Wang, D., Shang, Y.: A new active labeling method for deep learning. In: {Proceedings of the International Joint Conference on Neural Networks}. pp. 112--119. IEEE (2014)

\bibitem{Wang2017_CostEffectiveAL}
Wang, K., Zhang, D., Li, Y., Zhang, R., Lin, L.: Cost-effective active learning for deep image classification. IEEE Transactions on Circuits and Systems for Video Technology  \textbf{27}(12),  2591--2600 (2016)

\bibitem{welling2011bayesian}
Welling, M., Teh, Y.W.: {Bayesian} learning via stochastic gradient {Langevin} dynamics. In: Proceedings of the International Conference on Machine Learning. pp. 681--688 (2011)

\bibitem{Wu2022_surveyHumanInLoop}
Wu, X., Xiao, L., Sun, Y., Zhang, J., Ma, T., He, L.: A survey of human-in-the-loop for machine learning. Future Generation Computer Systems  \textbf{135},  364--381 (2022)

\bibitem{xie2018cooperative}
Xie, J., Lu, Y., Gao, R., Wu, Y.N.: Cooperative learning of energy-based model and latent variable model via {MCMC} teaching. In: Proceedings of the Annual AAAI Conference on Artificial Intelligence. vol.~32 (2018)

\bibitem{Xu2022energy}
Xu, Y., Xie, J., Zhao, T., Baker, C., Zhao, Y., Wu, Y.N.: Energy-based continuous inverse optimal control. IEEE Transactions on Neural Networks and Learning Systems pp. 1--15 (2022)

\bibitem{yang2023_medmnist}
Yang, J., Shi, R., Wei, D., Liu, Z., Zhao, L., Ke, B., Pfister, H., Ni, B.: {MedMNIST v2-A} large-scale lightweight benchmark for {2D} and {3D} biomedical image classification. Scientific Data  \textbf{10}(1), ~41 (2023)

\bibitem{Yang2021_JemPlusPlus}
Yang, X., Ji, S.: {JEM}++: Improved techniques for training {JEM}. In: {Proceedings of the IEEE/CVF International Conference on Computer Vision}. pp. 6494--6503 (2021)

\bibitem{Yang2023_sadaJEM}
Yang, X., Su, Q., Ji, S.: Towards bridging the performance gaps of joint energy-based models. In: {Proceedings of the IEEE/CVF Conference on Computer Vision and Pattern Recognition}. pp. 15732--15741 (2023)

\bibitem{Zhan2022_ActiveLearningSurvey}
Zhang, Z., Strubell, E., Hovy, E.: A survey of active learning for natural language processing. In: {Proceedings of the Conference on Empirical Methods in Natural Language Processing}. pp. 6166--6190. Association for Computational Linguistics (2022)

\end{thebibliography}




% LNCS Homepage, \url{http://www.springer.com/lncs}, last accessed 2023/10/25
% \end{thebibliography}
\end{document}